\pdfoutput=1

\documentclass[11pt]{article}

\usepackage{ACL2023}
\usepackage{times}
\usepackage{latexsym}

\usepackage[T1]{fontenc}
\usepackage[utf8]{inputenc}
\usepackage{microtype}
\usepackage{inconsolata}
\usepackage{algorithm}
\usepackage{algorithmic}
\usepackage{makecell}
\usepackage{multirow}
\usepackage{CJKutf8}
\usepackage{color}
\usepackage{pgfplots}
\usepackage[normalem]{ulem}
\usepackage{graphicx}
\usepackage{amsfonts}
\usepackage{amsmath,amssymb,bm}
\usepackage{times}
\usepackage{latexsym}
\usepackage{color,xcolor}
\usepackage{CJKutf8}
\usepackage{amsmath,amsfonts}
\usepackage{algorithmic}
\usepackage{array}
\usepackage[caption=false,font=normalsize,
   labelfont=sf,textfont=sf]{subfig}
\usepackage{textcomp}
\usepackage{stfloats}
\usepackage{url}
\usepackage{verbatim}
\usepackage{graphicx}
\usepackage{balance}
\usepackage{inconsolata}
\usepackage{booktabs}
\usepackage{paralist}
\usepackage{colortbl}
\usepackage{algorithm}
\usepackage{algorithmic}
\usepackage{makecell}
\usepackage{arydshln}
\usepackage{multirow}
\usepackage{color}
\usepackage{pgfplots}
\usepackage[normalem]{ulem}
\usepackage{graphicx}
\usepackage{amsfonts}
\usepackage{amsmath,amssymb,bm}
\usepackage{soul}
\usepackage{amsmath}
\usepackage{tabularx}
\usepackage{booktabs}
\usepackage{amssymb}
\usepackage{pifont}
\usepackage{bm}
\usepackage{multirow}
\usepackage[framemethod=tikz]{mdframed}
\usepackage{tikz,pgfplots}
\usepgfplotslibrary{groupplots}
\pgfplotsset{compat=1.13}

\definecolor{nmgray}{RGB}{229,229,229}
\definecolor{underlinegray}{RGB}{197,197,197}

\definecolor{introblue}{RGB}{0,176,240}
\definecolor{introgreen}{RGB}{0,203,134}
\definecolor{introgreen2}{RGB}{139,243,206}

\usepackage{adjustbox}

\usepackage[most]{tcolorbox}
\newtcolorbox{mybox}[2][]{
width=\columnwidth,
colback = nmgray!75!white, 
colframe = nmgray!75!white, 
boxsep=0pt,left=10pt,right=10pt,top=0pt,bottom=0pt,
fontupper=\linespread{0.9}\selectfont,
title=#2,#1}

\definecolor{remarkbg}{rgb}{0.927,1,1}
\definecolor{remarkborder}{gray}{0.15}
\colorlet{remarktitlebg}{remarkbg!20!black}

\newenvironment{remark}[1][]
  { 
 \begin{tcolorbox}
 [
    enhanced, 
    breakable,
    boxrule=0.5pt,
    arc=4pt,
    left=2pt,
    right=2pt,
    bottom=2pt,
    top=2pt, 
    rounded corners 
    ]{}
  \textbf{#1}
  \small \itshape
  }
  {
\end{tcolorbox} 
}

\newenvironment{prompt_yellow}[1][]
  { 
 \begin{tcolorbox}
 [
    enhanced, 
    breakable,
    boxrule=0.5pt,
    arc=3.8pt,
    left=1.8pt,
    right=1.8pt,
    bottom=2pt,
    top=2pt, 
    rounded corners,
    colback=yellow!10
    ]{}
  \textbf{#1}
  \small \itshape
  }
  {
\end{tcolorbox} 
}

\newcommand{\circleone}[1]{%
    \resizebox{!}{0.8em}{%
        \tikz[baseline=(char.base)]{
            \node[shape=circle, fill=black, inner sep=0.8pt, text=white] (char) {#1};
        }%
    }%
}

\newcommand{\circletwo}[1]{%
    \resizebox{!}{0.8em}{%
        \tikz[baseline=(char.base)]{
            \node[shape=circle, fill=black, inner sep=0.8pt, text=white] (char) {#1};
        }%
    }%
}

\newcommand{\circlethree}[1]{%
    \resizebox{!}{0.8em}{%
        \tikz[baseline=(char.base)]{
            \node[shape=circle, fill=black, inner sep=0.8pt, text=white] (char) {#1};
        }%
    }%
}

\newcommand{\circlefour}[1]{%
    \resizebox{!}{0.8em}{%
        \tikz[baseline=(char.base)]{
            \node[shape=circle, fill=black, inner sep=0.8pt, text=white] (char) {#1};
        }%
    }%
}

\usepackage{CJKutf8}
\usepackage{color}

\title{
Are Emotion and Rhetoric Neurons in LLM? Neuron Recognition and Adaptive Masking for Emotion-Rhetoric Prediction Steering
}

\author{
  Li Zheng\textsuperscript{\rm 1}, Xin Zhang\textsuperscript{\rm 2}, Shuyi He\textsuperscript{\rm 1}, Fei Li\textsuperscript{\rm 1}\thanks{     
    $\,$ Corresponding author.}, Chong Teng\textsuperscript{\rm 1}, \\ \textbf{Jiangming Yang\textsuperscript{\rm 2}, Donghong Ji\textsuperscript{\rm 1}\footnotemark[1], Zhuang Li
 \textsuperscript{\rm 3}}
  \\
  \textsuperscript{\rm 1}Key Laboratory of Aerospace Information Security and Trusted Computing, Ministry of \\ Education, School of Cyber Science and Engineering, Wuhan University, Wuhan, China\\
  \textsuperscript{\rm 2} Ant International
  \textsuperscript{\rm 3}School of Computing Technologies, RMIT University, Australia
  \\
\texttt{\{zhengli,heshuyi354,lifei\_csnlp,tengchong,dhji\}@whu.edu.cn} \\ 
    \texttt{\{evan.zx,jmyang\}@ant-intl.com},\ \texttt{zhuang.li@rmit.edu.au}}

\begin{document}
\begin{CJK}{UTF8}{gbsn}
\maketitle

\begin{abstract}

Accurate comprehension and controllable generation of emotion and rhetoric are pivotal for enhancing the reasoning capabilities of large language models (LLMs). 
Existing studies mostly rely on external optimizations, lacking in-depth exploration of internal representation mechanisms, thus failing to achieve fine-grained steering at the neuron level. 
A handful of works on neurons are confined to emotions, neglecting rhetoric neurons and their intrinsic connections. 
Traditional neuron masking also exhibits counterintuitive phenomena, making reliable verification of neuron functionality infeasible. 
To address these issues, we systematically investigate the neurons representation mechanisms and inherent associations of 6 emotion categories and 4 core rhetorical devices. 
We propose a neuron identification framework that integrates multi-dimensional screening, and design an adaptive masking method incorporating dynamic filtering, attenuation masking, and feedback optimization, enabling reliable causal validation of neuron functionality.
Through neuron regulation, we achieve directed induction of non-target sentences and enhancement of emotion tasks via rhetoric neurons. 
Experiments on 5 commonly used datasets validate the effectiveness of our method, providing a novel paradigm for the fine-grained steering of emotion and rhetoric expressions in LLMs.

\end{abstract}
\section{Introduction}

Emotion and rhetoric are fundamental components of human communication \cite{konstan2007rhetoric,fussell2014figurative,zheng2025improving,zheng2025enhancing}.
Emotions convey speakers' subjective attitudes and affective states, while rhetorical devices shape how meanings are intensified, softened, or implied.
As large language models (LLMs) are increasingly deployed in intelligent dialogue \cite{ou2024dialogbench,xu2024can,zheng2025ecqed}, creative writing \cite{wu2025one,zhong2024let}, and customer service \cite{su2025llm,li2025cusmer}, robust understanding and controllable expression of emotion and rhetoric become essential for improving user experience, safety, and reliability in real-world interactions \cite{hu2017toward}.

\begin{figure}[!t]
    \centering
    \includegraphics[width=\columnwidth]{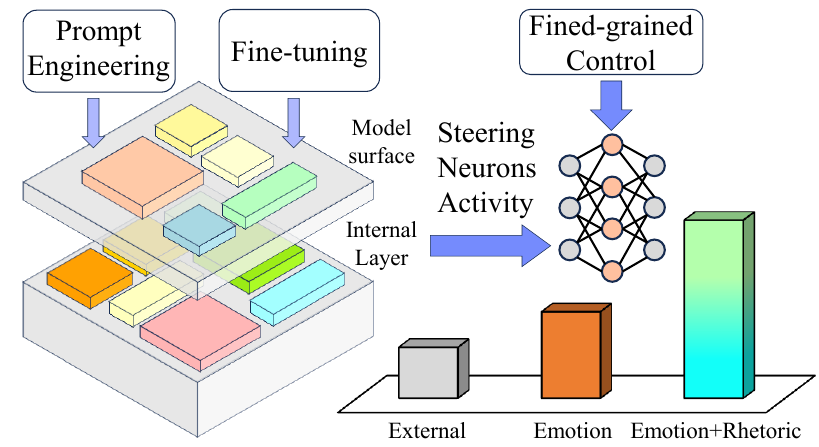}
    \caption{Emotion recognition accuracy comparison on DailyDialogue \cite{li2017dailydialog}: external optimization vs. emotion and rhetoric neurons steering. ``External'' denotes prompt engineering or fine-tuning. ``Emotion'' indicates the injection of emotion neurons. ``Emotion+Rhetoric'' denotes the injection of emotion neurons augmented with rhetoric neurons.}
    \label{fig:add}
    \vspace{-0.4cm}
\end{figure}

As illustrated in Figure~\ref{fig:add}, 
most existing approaches improve emotion- and rhetoric-related performance primarily through \emph{external} optimization, such as prompt engineering \cite{brown2020language,zheng2024reverse} or fine-tuning \cite{liu2024emollms}.
While effective at the output level, these methods provide limited causal insight into the internal representations that support emotion and rhetoric, and they often offer weak fine-grained controllability over specific affective or rhetorical attributes.
Meanwhile, recent interpretability studies have started to investigate \emph{emotion-selective neurons} \cite{lee2025large,di2025llamas}; however, two key gaps remain.
On the one hand, the internal mechanisms of \emph{rhetorical} representations, and more importantly, the relationship between emotion and rhetoric inside LLMs, are still underexplored, leaving unclear whether and how rhetorical signals modulate emotional representations.
On the other hand, widely used neuron-level \emph{function verification} practices (e.g., forced-zero ablation or mean substitution) can behave unreliably on these tasks: masking neurons that appear highly related to a target label does not necessarily lead to the expected performance degradation, and can even yield counterintuitive changes.
This raises a practical challenge: without a reliable intervention scheme, it is difficult to make trustworthy causal claims about neuron functionality, let alone use such neurons for controllable manipulation.

Motivated by these gaps, we systematically study neuron-level representations of emotion and rhetoric, as well as their interactions, covering 6 emotion categories (happiness (hap), sadness (sad), anger (ang), fear (fea), surprise (sup), and disgust (dig)) and 4 rhetorical devices (metaphor (met), hyperbole (hpy), humor (hum), and sarcasm (sar)).
Specifically, we ask:
\begin{prompt_yellow}
Can we reliably localize neurons that are selective to specific emotion and rhetoric labels, verify their causal roles through stable masking interventions, and leverage them for controllable steering, including probing potential emotion--rhetoric interactions?
\end{prompt_yellow}

To answer this question, we develop a neuron-level analysis and intervention framework.
\textbf{First}, we identify candidate emotion and rhetoric neurons by combining activation frequency statistics, probability normalization, and entropy-based selectivity filtering, and we characterize their layer-wise distributions.
\textbf{Second}, to address the instability and counterintuitive behaviors of traditional masking, we propose an \emph{adaptive masking} strategy that integrates dynamic neuron selection, attention-based masking, and feedback-driven adjustment.
This design aims to produce a consistent decline in task performance when masking truly relevant neurons, improving the reliability of causal verification.
\textbf{Third}, we demonstrate controllable manipulation by intervening on the identified neuron sets to steer predictions from non-target to target emotion/rhetoric categories.
Furthermore, we study cross-signal interactions by injecting rhetoric-neuron signals into emotion recognition, testing whether rhetorical representations can assist emotional discrimination.

We conduct extensive experiments on five widely used emotion and rhetoric datasets, namely DailyDialogue \cite{li2017dailydialog}, HYPO \cite{troiano2018computational}, TroFi \cite{birke2006clustering}, IAC-v2 \cite{oraby2016creating}, and ColBERT \cite{annamoradnejad2024colbert}, from which we derive the following insights.
\circleone{1} Emotion and rhetoric neurons tend to exhibit stronger activation in upper layers.
\circletwo{2} Rhetoric-neuron activations can improve the separability of emotional features, benefiting emotion recognition.
\circlethree{3} The proposed adaptive masking yields a stable performance decline after masking, enabling more reliable function verification than standard ablation/substitution baselines.
\circlefour{4} By manipulating the identified neurons, predictions for non-target emotion/rhetoric inputs can be directionally induced toward target categories.

Our main contributions are as follows:
\begin{itemize}
\item 

We conduct the first systematic investigation into the foundations of rhetoric neurons and confirm the auxiliary role of rhetoric in emotion recognition.

\item 
We propose an adaptive neuron masking method that enables reliable causal verification of neuron functions.

\item 
We achieve controllable manipulation of emotion and rhetoric neurons, offering a novel pathway for the fine-grained steering of emotional and rhetorical expressions in LLMs.

\end{itemize}

\section{Observation and Key Intuition}

Most existing approaches improve emotion- and rhetoric-related performance primarily via output-level optimization (e.g., prompting or fine-tuning), which provides limited neuron-level causal evidence and offers weak fine-grained controllability.
Moreover, when we attempt to verify neuron functionality using standard masking interventions (e.g., forced-zero or mean substitution), we observe counterintuitive behaviors on emotion and rhetoric tasks (Figure~\ref{fig:intro}), making causal verification unreliable in this setting.
Motivated by these observations, we develop an intervention-centric framework that localizes label-selective neurons and employs more stable masking and activation-based interventions for verification and steering.

\definecolor{neongreen}{HTML}{40E0D0}   
\definecolor{neonorange}{HTML}{FF8C00}  

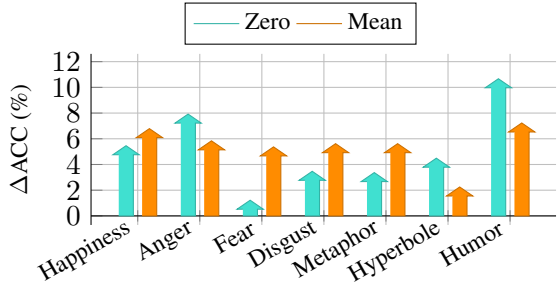
\begin{figure}
\begin{tikzpicture}
\begin{axis}[
    width=\columnwidth,
    height=0.5\columnwidth,
    ymin=-0.1, ymax=12,   
    ytick={0,2,4,6,8,10,12},
    ylabel={$\Delta$ACC (\%)},             
    ylabel style={font=\footnotesize},          
    ymajorgrids=true,
    xmajorgrids=true,
    axis x line=middle,   
    axis line style={-},
    axis y line*=left,
    xtick={0.35, 1.20, 2.05, 2.90, 3.75, 4.60, 5.45},
    xticklabels={Happiness, Anger, Fear, Disgust, Metaphor, Hyperbole, Humor},
    xticklabel style={rotate=30, anchor=east, font=\small},
    enlargelimits=0.05,
    legend style={font=\small, at={(0.7,1.3)}, legend columns=2, anchor=north east},
]

\newcommand{\fullarrow}[4]{
    \pgfmathsetmacro{\w}{#4}
    \pgfmathsetmacro{\ah}{0.7}   

    \addplot[
        draw=#3!80!black,
        line width=0.4pt,
        fill=#3
    ] coordinates {
        (#1-\w,0)
        (#1+\w,0)
        (#1+\w,#2-\ah)
        (#1+0.18,#2-\ah)
        (#1,#2)
        (#1-0.18,#2-\ah)
        (#1-\w,#2-\ah)
    };
}

\pgfmathsetmacro{\dx}{0.16}

\fullarrow{0.35-\dx}{5.46}{neongreen}{0.1}  
\fullarrow{0.35+\dx}{6.79}{neonorange}{0.1} 

\fullarrow{1.2-\dx}{7.92}{neongreen}{0.1}
\fullarrow{1.2+\dx}{5.84}{neonorange}{0.1}

\fullarrow{2.05-\dx}{1.22}{neongreen}{0.1}
\fullarrow{2.05+\dx}{5.37}{neonorange}{0.1}

\fullarrow{2.90-\dx}{3.47}{neongreen}{0.1}
\fullarrow{2.90+\dx}{5.62}{neonorange}{0.1}

\fullarrow{3.75-\dx}{3.37}{neongreen}{0.1}
\fullarrow{3.75+\dx}{5.62}{neonorange}{0.1}

\fullarrow{4.60-\dx}{4.49}{neongreen}{0.1}
\fullarrow{4.60+\dx}{2.25}{neonorange}{0.1}

\fullarrow{5.45-\dx}{10.67}{neongreen}{0.1}
\fullarrow{5.45+\dx}{7.23}{neonorange}{0.1}

\addplot[neongreen, thick] coordinates {(0,0)};  
\addlegendentry{Zero}

\addplot[neonorange, thick] coordinates {(0,0)}; 
\addlegendentry{Mean}
\end{axis}
\end{tikzpicture}
\caption{\textbf{Observation}. Performance changes of emotion and rhetoric tasks under traditional target neuron masking methods.}
\label{fig:intro}
\vspace{-0.4cm}
\end{figure}

\subsection{Observation on Existing Limitations}

\noindent\textbf{Limited neuron-level insight and steering.}
Output-level optimization methods (e.g., prompt engineering and fine-tuning) are effective for improving task performance, but they do not directly provide causal evidence about which internal representations are responsible for emotion and rhetoric predictions, nor do they enable targeted neuron-level steering.
In addition, the potential interaction between emotion and rhetoric representations inside LLMs is underexplored, limiting opportunities to leverage rhetorical cues to assist emotion recognition and to build more controllable systems.
Existing methods \cite{kim2025exploring,yang2025mse} fail to recognize this core mechanism, and only optimize the model output through surface-level text features, resulting in two major problems.
One is the lack of controllability and synergism.
Precise modulation of emotional or rhetorical expressions cannot be achieved, and the failure to explore the correlation between emotion and rhetoric results in the absence of rhetorical cues for emotional recognition tasks.
The other is insufficient interpretability, such that performance improvements cannot be traced back to the causal mechanism of neuronal functioning.

\noindent\textbf{Unreliable verification under standard masking.}
Masking-based methods aim to identify task-relevant hidden features by intervening on internal representations~\cite{feng2024imo}, yet standard interventions can behave counterintuitively on emotion and rhetoric tasks.
As shown in Figure~\ref{fig:intro}, forcing selected neurons to zero or substituting them with a mean value does not consistently reduce accuracy and can even increase it.
We hypothesize that such behaviors are consistent with redundancy or redistribution effects (e.g., alternative pathways becoming more dominant under hard ablation), and that mean substitution may leave residual information that still supports inference.
These observations motivate a more stable intervention scheme for function verification.

\subsection{Key Intuition of Our Method}

\noindent\textbf{Manipulation of internal neurons.}
The emotion and rhetoric  neurons exhibit functional specificity and activation stability.
Specific neurons only respond strongly to a certain type of emotion or rhetoric, and this response pattern remains consistent across different scenarios. 
More importantly, a synergistic activation effect exists between the two neurons. 
The activation of rhetoric neurons will significantly enhance the response intensity of the corresponding emotion neurons, providing a basis for the associated regulation of emotion and rhetoric.
As illustrated in Figure \ref{fig:add}, the performance of relying solely on external optimization is significantly weaker than that of directly regulating neurons. Notably, co-regulating emotion and rhetoric neurons yields superior task performance.

\noindent\textbf{Adaptive masking methods.}
The counterintuitive phenomenon observed in traditional masking methods stems essentially from a mismatch between the masking strategies and the functional characteristics of neurons. 
Thus, the key to successful masking lies in identifying genuinely core neurons via activation discrepancy screening, followed by intervening in their functions through attenuation masking instead of complete ablation. 
Attenuation masking not only avoids triggering the model’s functional compensation mechanism but also precisely weakens the representational capacity of target neurons, leading to a steady decline in task accuracy and enabling reliable validation of their functions.

\begin{remark}[Takeaway.]
We provide a systematic study of rhetoric-selective neurons and their interaction with emotion-selective neurons.
We further propose an adaptive, attenuation-based masking scheme that yields more stable verification behavior than standard masking baselines, and demonstrate neuron-level steering of emotion and rhetoric predictions via activation interventions.
\end{remark}

\section{Methodology}

\begin{figure*}[!h]
    \centering
    \includegraphics[scale=0.4]{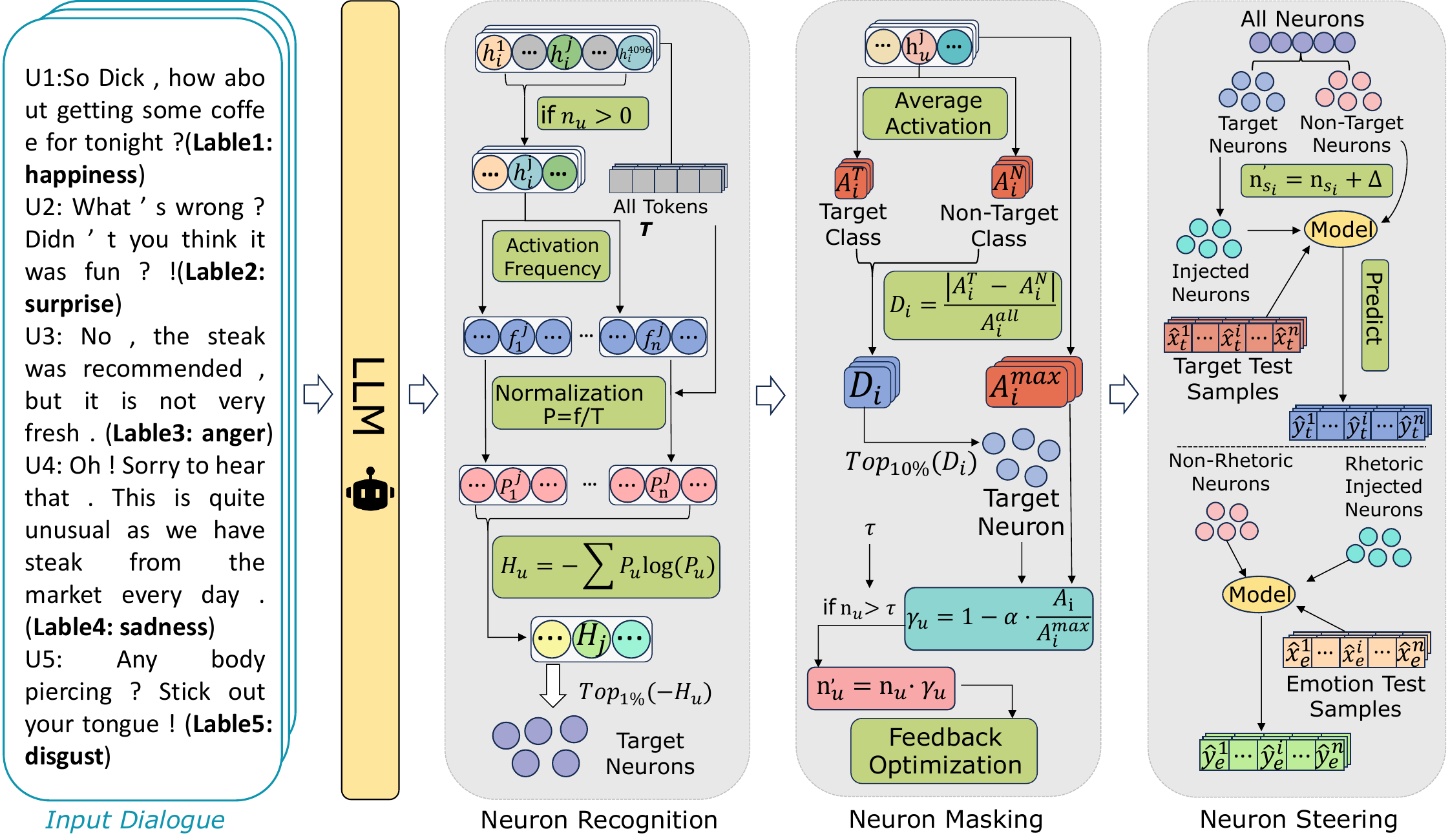}
    \caption{
    The overall architecture of our framework. $h_i^j$ denotes the representation of the $j$-th neuron in the $i$-th layer. The target class refers to a specific emotion or rhetoric category subject to manipulation.}
    \label{fig:model}
\end{figure*}

As illustrated in Figure \ref{fig:model}, our framework adopts the Llama-3.1-8B-Instruct as the research vehicle. 
This model is built on a Transformer decoder architecture, comprising 32 Transformer blocks where each block consists of a multi-head self-attention (MHA) module and a feed-forward network (FFN) module. 
Existing research \cite{lee2025large} has confirmed that FFN layers serve as the core representation region for semantic features, thus we focus on the neurons within FFN layers.

\subsection{Neuron Recognition}

\noindent\textbf{Neuron activation mechanism.}
After the MHA module, the FFN layer performs additional non-linear transformations on token features to further distill semantic information. 
\begin{equation}
\text{FFN}'(X_{\text{mid}}) = g(X_{\text{mid}} W_{\text{in}}) W_{\text{out}}
\end{equation}
where $X_{\text{mid}}$ denotes the output features of the MHA module, $W_{\text{in}}$ and $W_{\text{out}}$ represent the input and output weight matrices of the FFN layer, and $g(\cdot)$ is the activation function. 
For a neuron $u$ within the FFN layer, its activation condition is defined as:
\begin{equation}
n_u = \max(0, h_u)
\end{equation}
where $h_u$ is the output of neuron $u$ after linear transformation. Neuron $u$ is determined to be activated when $n_u > 0$.

\noindent\textbf{Recognizing neurons.}
We first calculate the activation frequency. 
For each neuron $n_u$ in the FFN layer of each model layer, we count its activation times when inputting sentences corresponding to a emotion label e or rhetorical label r, denoted as $f_{u,e}$ or $f_{u,r}$. 
Then, we perform activation probability normalization. 
Let \( T_e \) (or \( T_r \)) be the total number of tokens corresponding to emotion \( e \) or rhetoric \( r \). 
The activation probability \( P_{u,e} \) or \( P_{u,r} \) is given by:
\begin{equation}
\begin{split}
P_{u,e} = \frac{f_{u,e}}{T_e} , P_{u,r} = \frac{f_{u,r}}{T_r}
\end{split}
\end{equation}

Finally, we use entropy to measure the concentration of this probability distribution. 
A lower entropy indicates that the neuron’s responses are more concentrated on specific emotions or rhetorics. 
The entropy is calculated as follows:
\begin{equation}
\begin{split}
H_{e/r} = -\sum_{e/r \in E/R} P_{u,e/r} \log(P_{u,e/r})
\end{split}
\end{equation}
where E denotes the set of 6 emotion categories, and R denotes the set of 4 rhetorical categories. 
We select the top 1\% neurons with the lowest entropy as emotion or rhetoric neurons.

\subsection{Neuron Masking}

\noindent\textbf{Traditional masking methods.}
Traditional masking methods verify functionality by directly intervening in neuron activation values, encompassing two classical implementation paradigms. 
One is the forced zero method, which directly resets the activation value of the target neuron to 0, with the formula given as follows:
\begin{equation}
n_u' = 0
\end{equation}
where \( n_u' \) denotes the activation value of the neuron after masking.

The other is the mean substitution method, which replaces the original activation value of the target neuron with the global mean of the activation values of all non-target neurons in the FFN layer where the target neuron resides. 
\begin{equation}
n_u' = \frac{1}{N - M} \sum_{k \in \Omega} n_k
\end{equation}
where \( N \) is the total number of neurons in the FFN layer, \( M \) is the number of target neurons, \( \Omega \) represents the set of all non-target neurons in the layer, and \( n_k \) is the original activation value of the \( k \)-th non-target neuron in \( \Omega \).

\noindent\textbf{Adaptive masking methods.}
To address the counterintuitive phenomenon of traditional masking methods in emotion and rhetorical tasks, we propose an adaptive masking method integrating dynamic selection and feedback optimization. Throughout our experiments, the LLM parameters are kept fixed; we intervene only on FFN activations.
The feedback optimization (adjusting the selection criterion and $\alpha$) is performed on a held-out development split only; test sets are used once for final reporting without further updates. 

First, we collect neuron activation patterns. 
We feed the data into the model, and record the average activation value of each neuron \( i \) in each FFN layer for the two types of sentences, denoted as \( A_{i,\text{target} }\) (average activation for target sentences) and \( A_{i,\text{non-target}} \) (average activation for non-target sentences), respectively. 
On this basis, we define the activation difference \( D_i \) to quantify the response specificity of neuron \( i \):
\begin{equation}
D_i = \frac{|A_{i,target} - A_{i,non-target}|}{A_{all}} 
\end{equation}
where \( A_{\text{all}} \) denotes the average activation for all sentences. 
A larger \( D_i \) indicates that neuron \( i \) possesses a stronger ability to discriminate between target and non-target sentences.

Subsequently, we perform core neuron selection. 
By setting an activation threshold $\tau$ , we select the top 10\% of neurons ranked by \( D_i \) to construct a set \( S \) of target neurons that are significantly activated only in target sentences, ensuring that subsequent masking operations focus on critical neurons. 
Then we perform dynamic masking by applying attenuated masking to neurons in set \( S \). 
The activation value of the neuron after masking is as follows:
\begin{equation}
n_u' = n_u \times (1 - \alpha \times \frac{A_i}{A_i^{max}} )
\end{equation}
where \( \alpha \) denotes the attenuation coefficient, which is used to flexibly adjust the masking intensity.

To further ensure the reliability of the masking effect, we introduce a feedback optimization mechanism for iterative adjustment. 
The first step is accuracy monitoring. 
We calculate the task accuracy \( \text{Acc}_{\text{adapt}} \) after adaptive masking and conduct a quantitative comparison with the original accuracy \( \text{Acc}_{\text{origin}} \) before masking. 
The second step is dynamic parameter adjustment. 
If \( \text{Acc}_{\text{adapt}} \geq \text{Acc}_{\text{origin}} \), 
we increase the core neuron selection threshold $\tau$ and the attenuation coefficient \( \alpha \) to enhance the intensity of the masking operation.

\subsection{Neuron Steering}

The core objective of controllable neuronal manipulation is to achieve the controllable output of emotions and rhetorical devices by precisely regulating the activation states of emotion and rhetoric neurons. 
Concurrently, we investigate the intrinsic correlation between emotion and rhetoric, and validate the enhancing effect of rhetoric neurons on emotion recognition tasks.

\noindent\textbf{Rhetoric and emotion neuron steering.}
Based on the identified set of core target neurons, we extract the activation values of the neurons within this set to construct a functional vector \( \mathbf{V} = [n_{s_1}, n_{s_2}, ..., n_{s_k}] \). 
\( k \) denotes the number of neurons in the core neuron set, and \( n_{s_i} \) represents the original activation value of the \( i \)-th neuron in the set.
When a non-target-type sentence is input, we apply activation to all neurons in the core target neuron set within the FFN layer of the model to induce the model to generate a target-type output. 
\begin{equation}
n_{s_i}' = n_{s_i} + \beta \times \overline{n}_{s_i}
\end{equation}
where \( n_{s_i}' \) denotes the modulated activation value of the neuron \( s_i \), \( n_{s_i} \) is the original activation value of this neuron when a non-target sentence is input, \( \overline{n}_{si} \) represents the average activation value of the neuron \( s_i \) under the input scenario of target sentences, and \( \beta \) is the activation intensity coefficient.

Finally, we validate the manipulation effect by comparing the model’s classification results of non-target sentences before and after manipulation. 
If the model classifies sentences originally predicted as the non-target category into the target category after manipulation, it demonstrates the effectiveness of the neuronal manipulation.

\noindent\textbf{Rhetoric neurons assisted emotion recognition.} 
To validate the auxiliary enhancement effect of rhetoric on emotion recognition, we take metaphor as an example for illustration.
First, we extract metaphor-selective neurons in a given FFN layer (width $d$), represented as an index set $\mathcal{I}_{\text{meta}} \subseteq \{1,\dots,d\}$.
For each neuron index $i \in \mathcal{I}_{\text{meta}}$, we record its average activation under metaphorical inputs, calculated as follows:
\begin{equation}
\overline{a}_{i,\text{meta}} = \frac{1}{T_{\text{meta}}} \sum_{t=1}^{T_{\text{meta}}} a_{i,t},
\end{equation}
where $a_{i,t}$ denotes the activation of the $i$-th FFN neuron in the $t$-th metaphor sample,
and $T_{\text{meta}}$ is the number of metaphor samples. We store these values as a metaphor
feature library $\mathcal{F}_{\text{meta}}=\{\overline{a}_{i,\text{meta}}\}_{i\in\mathcal{I}_{\text{meta}}}$.

Then, when performing the emotion recognition task, we extract emotion-selective neurons as another index set
$\mathcal{I}_{\text{emo}} \subseteq \{1,\dots,d\}$ in the same FFN layer, and denote the
activation of neuron $i$ as $a_i$. We fuse neuron activations by element-wise injection.
The final fused activation value of the emotion neuron is calculated as follows:
\begin{equation}
a_{i,\text{joint}} = a_i + \omega \cdot \overline{a}_{i,\text{meta}}, \quad \forall i \in \mathcal{I}_{\text{emo}},
\end{equation}
where $\omega \in [0,1]$ controls the injection strength. Note that the index $i$ refers to the
same hidden unit in the same FFN layer for both $a_i$ and $\overline{a}_{i,\text{meta}}$.

\begin{figure*}[!t]
    \centering

    \makebox[\textwidth][l]{
    \begin{minipage}[t]{0.32\textwidth}
        \centering
        \includegraphics[width=\linewidth]{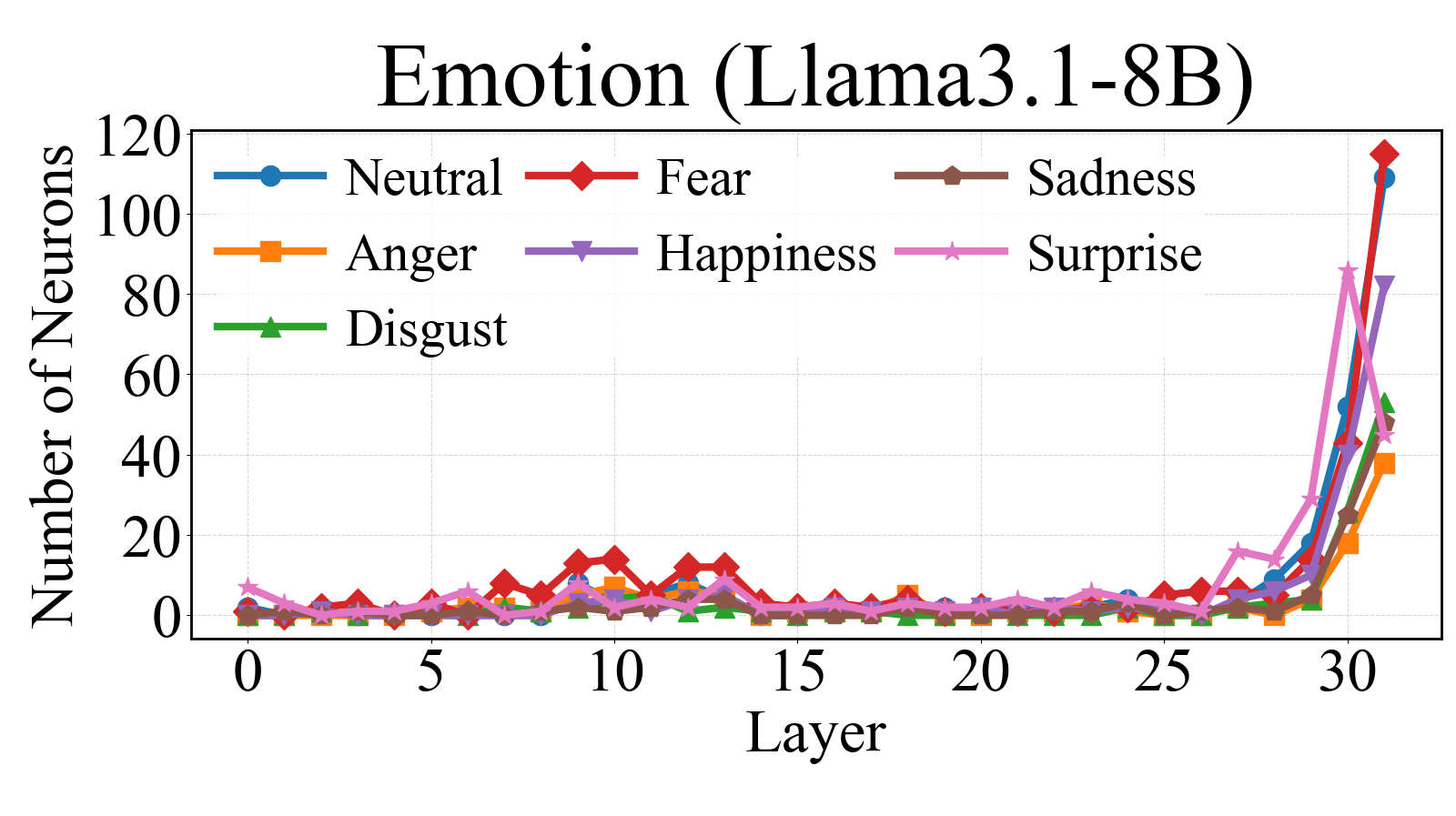}
        \vspace{0.1cm}
        \label{fig:sub1}
    \end{minipage}
    \hfill
    \begin{minipage}[t]{0.32\textwidth}
        \centering
        \includegraphics[width=\linewidth]{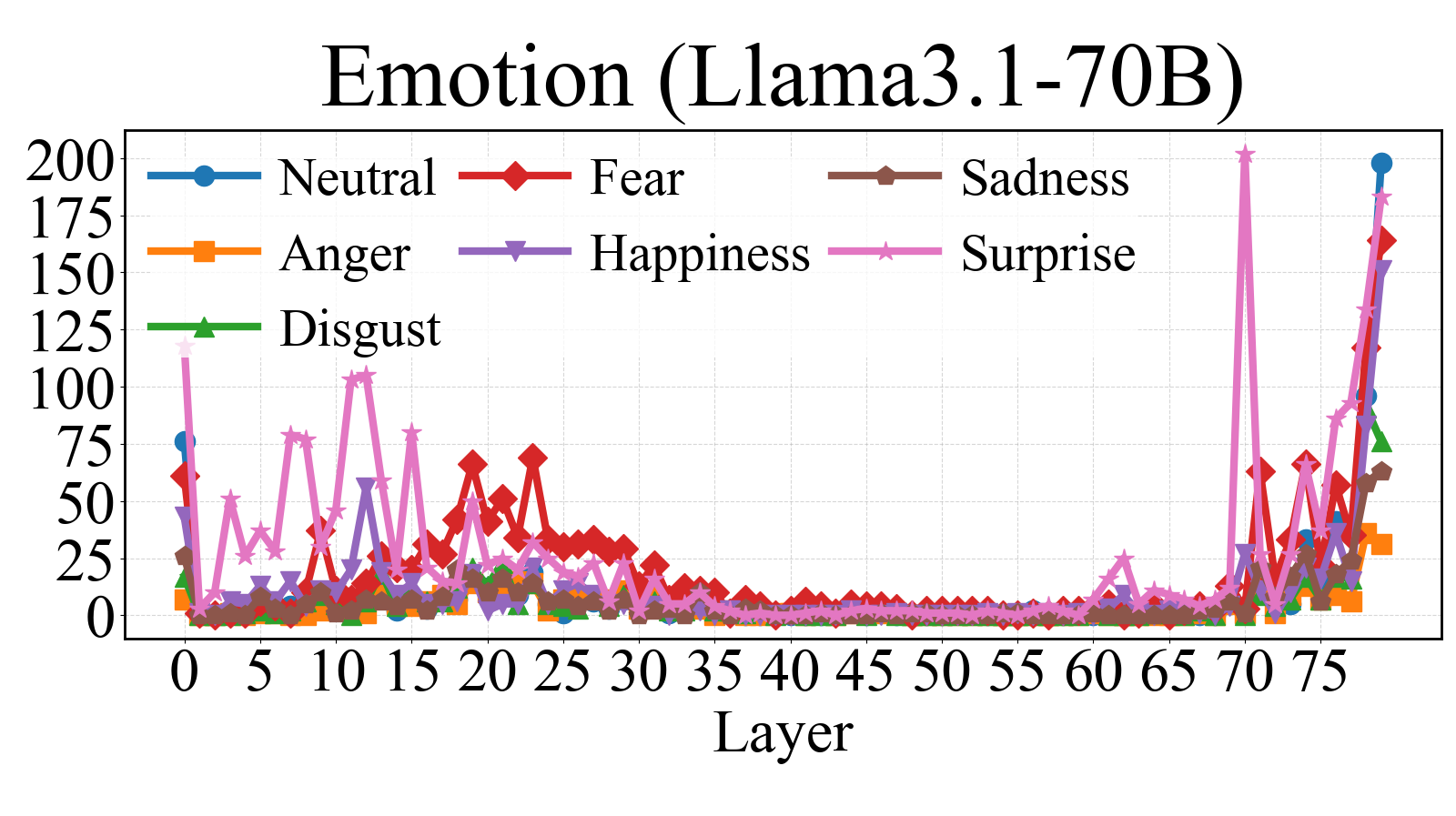}
        \vspace{0.1cm}
        \label{fig:sub2}
    \end{minipage}
    \hfill
    \begin{minipage}[t]{0.32\textwidth}
        \centering
        \includegraphics[width=\linewidth]{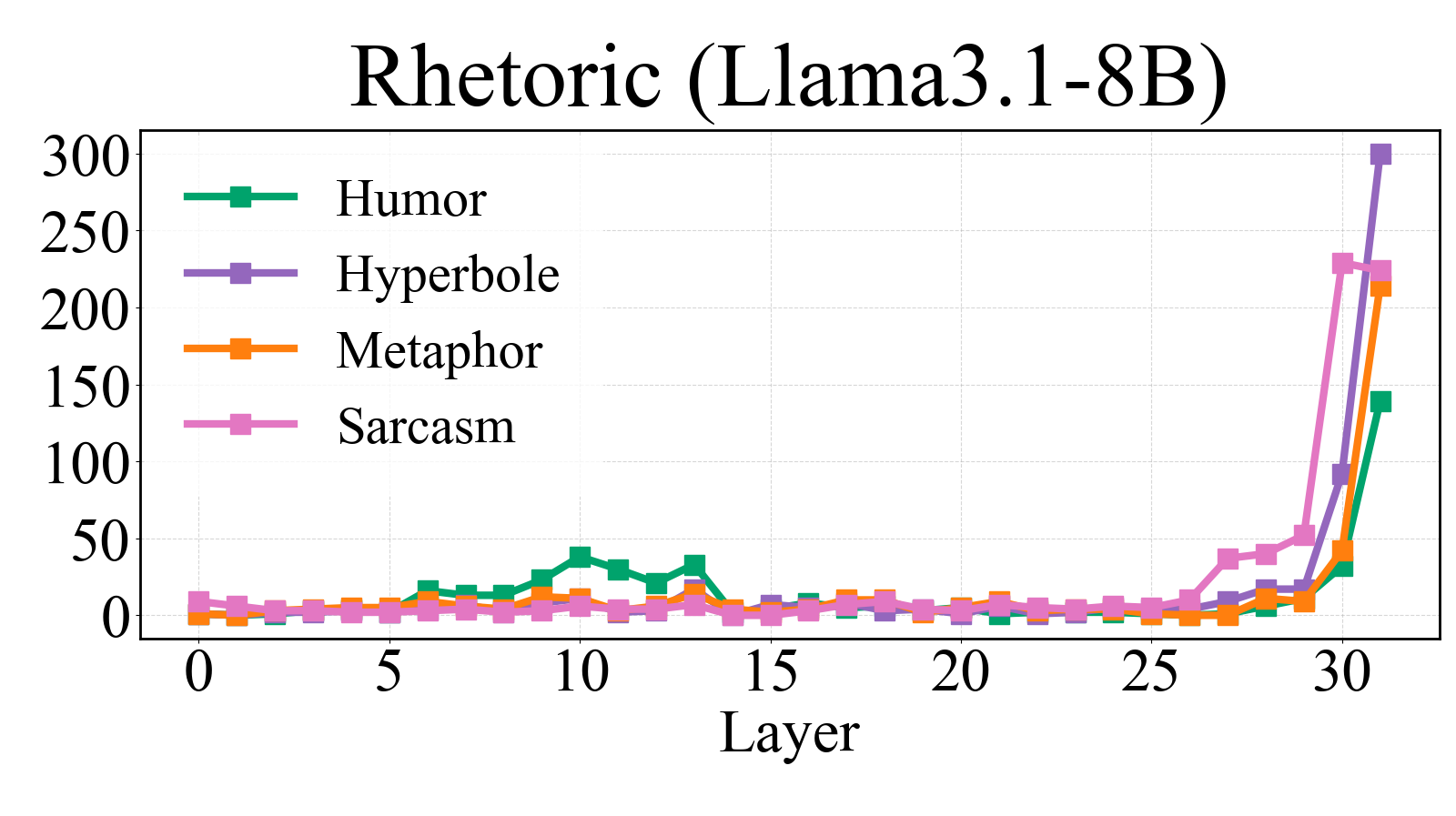}
        \vspace{0.1cm}
        \label{fig:sub3}
    \end{minipage}
    }

    \vspace{-6mm}
    \caption{Distribution of emotion and rhetoric neurons.}
    \label{fig:number}
\end{figure*}

\begin{table*}[htbp]
  \centering
  \resizebox{\textwidth}{!}{
  \begin{tabular}{lcccccccccc}
    \toprule
    \multirow{2}{*}{\textbf{Method}} & \multicolumn{6}{c}{\textbf{Emotion}} & \multicolumn{4}{c}{\textbf{Rhetoric}} \\
    \cmidrule(lr){2-7} \cmidrule(lr){8-11}
    & Happiness & Sadness & Anger & Fear & Surprise & Disgust & Metaphor & Hyperbole & Humor & Sarcasm \\
    \midrule
    \rowcolor{gray!15} Zero     & +5.46 & +4.32 & +7.92 & +1.22 & -2.85 & +3.47 & +3.37 & +4.49 & +10.67 & -4.17 \\
    Mean     & +6.79 & -1.13 & +5.84 & +5.37 & -3.59 & +5.62 & +5.62 & +2.25 & +7.23 & -5.78 \\
    \rowcolor{gray!15} 
    Adaptive & \textbf{-9.25} & \textbf{-8.63} & \textbf{-10.14} & \textbf{-7.85} & \textbf{-11.42} & \textbf{-14.26} & \textbf{-7.29} & \textbf{-13.48} & \textbf{-4.28 }& \textbf{-10.61} \\
    \midrule
      \multicolumn{11}{l}{\textbf{\textit{Cross-Dataset}}} \\
      \rowcolor{gray!15}  Zero& +3.24 &+5.11&+4.39& +2.37&-3.72& +1.68& +5.64& +5.28 &+9.31&-3.95\\
      Mean & +4.61& -1.18&+4.79&+5.28&-4.11&+4.65& +6.14& +3.71& +7.58& -5.73\\
      \rowcolor{gray!15} Adaptive & \textbf{-7.31} & \textbf{-6.83}& \textbf{-10.25} & \textbf{-6.37}&\textbf{ -12.39}& \textbf{-11.65}& \textbf{-6.79}& \textbf{-14.27}& \textbf{-5.96}& \textbf{-11.63}\\
    \bottomrule
  \end{tabular}
  } 
  \caption{Comparison results of different masking methods (accuracy change $\Delta\text{ACC (\%)}$).}
  \label{tab:mask}
\end{table*}

\section{Experiments}
\subsection{Experimental Setting}

We conduct evaluations on five widely used emotion and rhetoric datasets, namely DailyDialogue (emotion) \cite{li2017dailydialog}, TroFi (metaphor) \cite{birke2006clustering},  HYPO (hyperbole) \cite{troiano2018computational}, ColBERT (humor) \cite{annamoradnejad2024colbert}, and IAC-v2 (sarcasm) \cite{oraby2016creating}. 
In terms of evaluation metrics, we adopt Acc to assess the model’s performance.

\subsection{Localization and Distribution of Emotion and Rhetoric Neurons}

In Figure~\ref{fig:number}, we visualize the layer-wise distribution of the \emph{localized} emotion- and rhetoric-selective neurons and analyze their activation patterns across model layers.
For Llama-3.1-8B, both emotion and rhetoric neurons exhibit distinct top-layer aggregation: activation remains consistently low across the first 25 layers, surges sharply, and peaks at the top layers. 
This distribution suggests the model relies more on top-layer semantic integration to represent and process emotions and rhetoric. 
In contrast, emotion neurons in Llama-3.1-70B show dual-end concentration, with high activation in both top and bottom layers. This indicates that as model scale increases, emotional representation is no longer confined to upper-layer semantic integration, bottom layers also participate in early capture and initial processing of emotion information.

\subsection{Impact of Different Masking Methods}

\paragraph{Setting.} To validate the functional relevance of the localized target neurons, we keep the LLM parameters frozen and evaluate task accuracy under activation-level masking interventions.
We optimize the adaptive masking policy (including the core-neuron selection threshold and attenuation coefficient $\alpha$) on the development split of the original dataset, and report results on the original test set and five cross-datasets (IEMOCAP \cite{busso2008iemocap}, LCC \cite{mohler2016introducing}, HYPO-L \cite{zhang2022mover}, FunLines \cite{hossain2020stimulating}, MUStARD \cite{castro2019towards}).
We compare three masking methods: Zero masking (forced zeroing), Mean masking (mean substitution), and our adaptive masking.

\paragraph{Analysis.}
Experimental results in Table \ref{tab:mask} show that adaptive masking consistently reduces accuracy on both the original test set and cross-datasets, providing evidence that the localized neuron sets are causally relevant to model decisions under our intervention scheme.
 In contrast, Zero masking exhibits a counterintuitive trend where accuracy increases post-masking, which we conjecture stems from functionally complementary neuron clusters within the model. Mean masking fails to induce notable accuracy drops in most tasks: mean substitution does not truly disrupt core neurons’ specific functional representations, leaving residual semantic encoding intact and enabling the model to infer via residual features.

\subsection{Rhetoric Neurons Assisted Emotion Recognition}

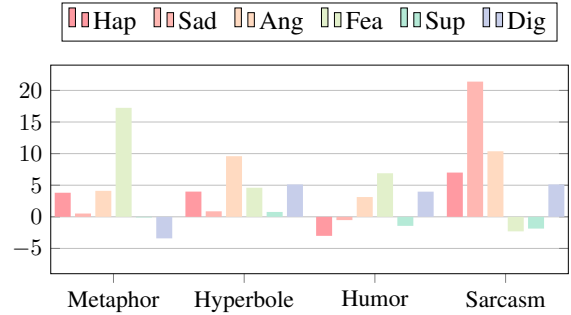
\begin{figure}[!t]
\resizebox{0.49\textwidth}{!}{ 
\begin{tabular}{c}

\hspace*{-0.2cm}
\begin{tikzpicture}[font=\normalsize]
\begin{axis}[
    ybar,
    width=10cm,
    height=5cm,
    ymin=-5,
    ymax=20,
    ytick distance=5,
    xtick pos=bottom,
    symbolic x coords={Metaphor, Hyperbole, Humor, Sarcasm},
    xticklabel style={font=\normalsize},
    xtick=data,
    bar width=0.26cm,
    enlargelimits=0.16,
    legend style={
        at={(0.5,1.3)},
        anchor=north,
        legend columns=6,
        font=\large,
        /tikz/every even column/.append style={column sep=6pt}
    },
    tick label style={font=\normalsize},
    ymajorgrids,
    grid style={gray!50, very thin},
]


\addplot[fill={rgb,255:red,255;green,154;blue,162}, draw=none] coordinates {
    (Metaphor, 3.81)
    (Hyperbole, 3.99)
    (Humor, -3.01)
    (Sarcasm, 7)
};

\addplot[fill={rgb,255:red,255;green,183;blue,178}, draw=none] coordinates {
    (Metaphor, 0.52)
    (Hyperbole, 0.87)
    (Humor, -0.52)
    (Sarcasm, 21.39)
};

\addplot[fill={rgb,255:red,255;green,218;blue,193}, draw=none] coordinates {
    (Metaphor, 4.11)
    (Hyperbole, 9.59)
    (Humor, 3.13)
    (Sarcasm, 10.37)
};

\addplot[fill={rgb,255:red,226;green,240;blue,203}, draw=none] coordinates {
    (Metaphor, 17.24)
    (Hyperbole, 4.6)
    (Humor, 6.89)
    (Sarcasm, -2.3)
};

\addplot[fill={rgb,255:red,181;green,234;blue,215}, draw=none] coordinates {
    (Metaphor, -0.1)
    (Hyperbole, 0.77)
    (Humor, -1.43)
    (Sarcasm, -1.86)
};

\addplot[fill={rgb,255:red,199;green,206;blue,234}, draw=none] coordinates {
    (Metaphor, -3.41)
    (Hyperbole, 5.12)
    (Humor, 3.98)
    (Sarcasm, 5.12)
};

\legend{Hap, Sad, Ang, Fea, Sup, Dig}

\end{axis}
\end{tikzpicture}

\end{tabular}
}
\caption{Experimental results of injecting rhetoric neurons into emotion recognition task.}
\label{fig:help}
\end{figure}

We inject rhetoric neurons into emotion texts to investigate their auxiliary enhancement on emotion recognition, with results in Figure \ref{fig:help}. 
Four typical rhetorical types exert positive effects on most emotion categories, verifying the potential of rhetoric neurons to boost emotion recognition performance. 
Metaphor neurons yield particularly prominent improvements for fear, as metaphors strengthen  perception of emotions via concrete expressions. 
Hyperbole neurons exert positive effects across all emotions, aligning with their inherent trait of amplifying emotion intensity. 
Humor neurons negatively impact emotions like happiness and surprise, as humor’s playful tone dilutes their original emotion concentration, but still benefit some emotions (e.g., anger, fear). 
Sarcasm neurons most significantly promote sadness, attributed to expressive compatibility between sarcasm’s implicit criticism and sadness’s restrained traits, enhancing the model’s recognition of this emotion.

\subsection{Impact of Masking Across Different Layers}

\definecolor{color1}{RGB}{60,155,201}
\definecolor{color2}{RGB}{118,203,180}
\definecolor{color3}{RGB}{248,132,85}
\definecolor{color4}{RGB}{252,117,123}
\definecolor{color5}{RGB}{253,202,147}
\definecolor{color6}{RGB}{252,229,155}

\definecolor{color7}{RGB}{236,128,76}
\definecolor{color8}{RGB}{119,214,221}
\definecolor{color9}{RGB}{167,181,253}
\definecolor{color10}{RGB}{255,57,56}

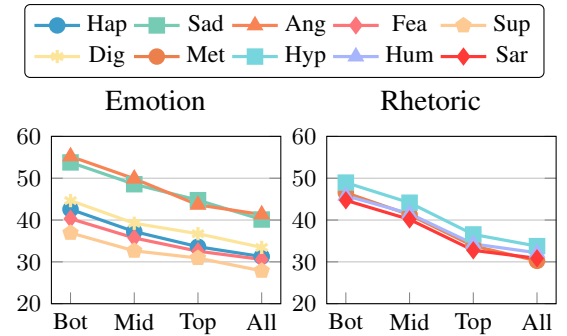
\begin{figure}[!t]
\centering

\begin{tikzpicture}[font=\small]
\begin{axis}[
    hide axis,
    xmin=0, xmax=1, ymin=0, ymax=1, 
    legend columns=5,               
    legend style={
        at={(0.5,0.1)},             
        anchor=south,
        draw,                       
        fill=white,                 
        rounded corners=2pt
    },
]

\addlegendimage{color=color1, mark=*, mark size=2.5pt, line width=1.2pt} \addlegendentry{Hap}
\addlegendimage{color=color2, mark=square*, mark size=2.5pt, line width=1.2pt} \addlegendentry{Sad}
\addlegendimage{color=color3, mark=triangle*, mark size=2.5pt, line width=1.2pt} \addlegendentry{Ang}
\addlegendimage{color=color4, mark=diamond*, mark size=2.5pt, line width=1.2pt} \addlegendentry{Fea}
\addlegendimage{color=color5, mark=pentagon*, mark size=2.5pt, line width=1.2pt} \addlegendentry{Sup}
\addlegendimage{color=color6, mark=asterisk, mark size=2.5pt, line width=1.2pt} \addlegendentry{Dig}
\addlegendimage{color=color7, mark=*, mark size=2.5pt, line width=1.2pt} \addlegendentry{Met}
\addlegendimage{color=color8, mark=square*, mark size=2.5pt, line width=1.2pt} \addlegendentry{Hyp}
\addlegendimage{color=color9, mark=triangle*, mark size=2.5pt, line width=1.2pt} \addlegendentry{Hum}
\addlegendimage{color=color10, mark=diamond*, mark size=2.5pt, line width=1.2pt} \addlegendentry{Sar}

\end{axis}
\end{tikzpicture}

\setlength{\tabcolsep}{2pt}
\begin{tabular}{cc}
\begin{tikzpicture}[font=\normalsize]
\begin{axis}[
    title={Emotion},
    title style={font=\normalsize, xshift=-4pt},
    height=3.8cm,
    width=4.6cm,  
    symbolic x coords={Bot, Mid, Top, All},
    xticklabel style={font=\small},
    xtick=data,
    xtick pos = bottom,
    ymin=20,
    ymax=60,
    yticklabel style={font=\small},
    ytick distance = 10,
    ymajorgrids,
    grid style={gray!50, very thin},
    legend columns=3,                     
    legend style={at={(0.5,1.6)},       
                  anchor=north,           
                  font=\scriptsize,
                  cells={anchor=west},    
                 }, 
    mark options={solid}
]

\addplot[color=color1, mark=*, mark size=2.5pt, line width=1.2pt] coordinates {
    (Bot, 42.56) 
    (Mid, 37.24)
    (Top, 33.62)
    (All, 31.26)
};

\addplot[color=color2, mark=square*, mark size=2.5pt, line width=1.2pt] coordinates {
    (Bot, 53.77) 
    (Mid, 48.56)
    (Top, 44.75)
    (All, 40.12)
};

\addplot[color=color3, mark=triangle*, mark size=2.5pt, line width=1.2pt] coordinates {
    (Bot, 55.21) 
    (Mid, 49.88)
    (Top, 43.62)
    (All, 41.32)
};

\addplot[color=color4, mark=diamond*, mark size=2.5pt, line width=1.2pt] coordinates {
    (Bot, 40.32) 
    (Mid, 35.71)
    (Top, 32.55)
    (All, 30.58)
};

\addplot[color=color5, mark=pentagon*, mark size=2.5pt, line width=1.2pt] coordinates {
    (Bot, 36.93) 
    (Mid, 32.64)
    (Top, 30.91)
    (All, 27.85)
};

\addplot[color=color6, mark=asterisk, mark size=2.5pt, line width=1.2pt] coordinates {
    (Bot, 44.65) 
    (Mid, 39.23)
    (Top, 36.73)
    (All, 33.46)
};

\end{axis}
\end{tikzpicture}
&
\hspace*{-0.3cm}
\begin{tikzpicture}[font=\normalsize]
\begin{axis}[
    title={Rhetoric},
    title style={font=\normalsize, xshift=-4pt},
    height=3.8cm,
    width=4.6cm,  
    symbolic x coords={Bot, Mid, Top, All},
    xticklabel style={font=\small},
    xtick=data,
    xtick pos = bottom,
    xticklabel style={font=\small},
    ymin=20,
    ymax=60,
    yticklabel style={font=\small},
    ymajorgrids,
    grid style={gray!50, very thin},
    ytick distance = 10,
    legend style={at={(0.5,1.48)},       
                  anchor=north,           
                  font=\scriptsize,
                  legend columns=4,     
                  cells={anchor=west},    
                 },  
    mark options={solid}
]

\addplot[color=color7, mark=*, mark size=2.5pt,line width=1.2pt] coordinates {
    (Bot, 46.53) 
    (Mid, 41.27)
    (Top, 33.82)
    (All, 30.24)
};

\addplot[color=color8, mark=square*,mark size=2.5pt, line width=1.2pt] coordinates {
    (Bot, 48.95) 
    (Mid, 44.14)
    (Top, 36.56)
    (All, 33.76)
};

\addplot[color=color9, mark=triangle*,mark size=2.5pt, line width=1.2pt] coordinates {
    (Bot, 45.82) 
    (Mid, 41.53)
    (Top, 34.29)
    (All, 32.17)
};

\addplot[color=color10, mark=diamond*,mark size=2.5pt, line width=1.2pt] coordinates {
    (Bot, 44.71) 
    (Mid, 40.14)
    (Top, 32.75)
    (All, 30.84)
};

\end{axis}
\end{tikzpicture}
\end{tabular}
\caption{Comparison results of different layers of masking.}
\label{figure:layer}
\vspace{-0.4cm}
\end{figure}

In Figure \ref{figure:layer}, we explore the impact of neuron masking across different layers on task prediction outcomes. 
We find that all layer masking induces the most significant performance degradation across all emotion and rhetoric tasks. 
This indicates that the functional representations of emotion and rhetoric neurons are not confined to a single layer but instead rely on cross-layer synergistic interactions.
Beyond all layer masking, top layer masking exerts the most pronounced performance impairment on emotion and rhetoric tasks, outperforming mid and bottom layer masking in terms of prediction accuracy reduction. 
This observation aligns with the distribution characteristic that emotion and rhetoric neurons exhibit higher activation intensity in the top layers. 
Since the top layers aggregate a larger number of emotion and rhetoric neurons, masking these layers deprives the model of effective semantic encoding support for emotion and rhetoric. 
This further validates the hierarchical distribution characteristics and functional relevance of emotion and rhetoric neurons in the model’s architecture.

\subsection{Neuron-level Steering of Emotion and Rhetoric Predictions}

\definecolor{before}{HTML}{AEE6E1}
\definecolor{after}{HTML}{FFABB7}
\begin{figure}[!t]
\centering

\begin{tikzpicture}[baseline]
    \node[draw, rounded corners=0pt, inner sep=1.5pt] (legend) {
     
        \begin{tikzpicture}[baseline]
            \node[draw, fill=before, minimum width=6pt, minimum height=6pt] at (0,0) {};
            \node[anchor=west] at (0.2,0) {Before};

            \node[draw, fill=after, minimum width=6pt, minimum height=6pt] at (1.6,0) {};
            \node[anchor=west] at (1.8,0) {After};
        \end{tikzpicture}
    };
\end{tikzpicture}

\vspace{1pt}

\setlength{\tabcolsep}{3pt}
\begin{tabular}{cc}

\begin{minipage}{4cm}
    \centering
    \includegraphics[width=\linewidth]{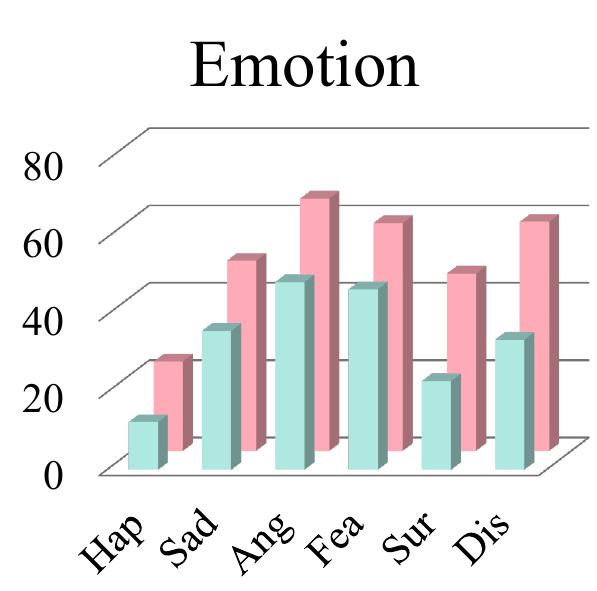}
\end{minipage}
&

\begin{minipage}{4cm}
    \centering
    \includegraphics[width=\linewidth]{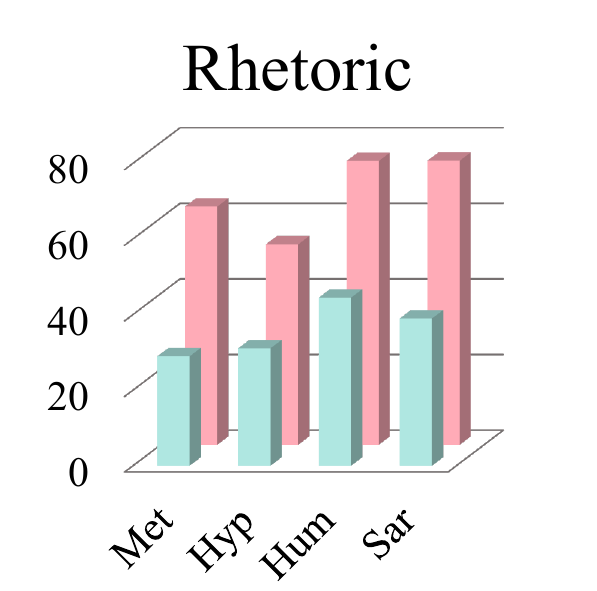}
\end{minipage}

\end{tabular}

\caption{Comparison results before and after neurons manipulation.}
\label{figure:control}
\vspace{-0.4cm}
\end{figure}
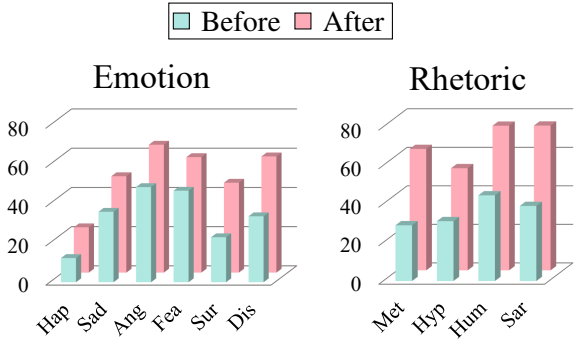

We validate the controllable manipulation capability of emotion and rhetoric neurons in Figure \ref{figure:control}. 
Specifically, we inject the functional features of target neurons into non-target sentences, thereby inducing the model to convert its predictions of non-target types into target types.
The manipulation efficacy is quantified as the ratio of samples predicted as the target class to the total number of samples in the dataset. 
A higher ratio indicates that more non-target instances are induced to be classified as the target class, thus reflecting a stronger controllable manipulation capability. 
When manipulating emotion neurons, the coverage rate of all emotion categories is significantly improved after the injection of target neuron features. 
This indicates that modulation of the activation states of emotion neurons can effectively enhance the model’s perception of target emotion features. 
Similarly, the controllability verification yields excellent results in the rhetoric neuron manipulation task. 
This demonstrates that the injection of functional features of rhetoric neurons can precisely guide the model to capture the semantic expression patterns of target rhetorical devices.
It also successfully induces non-target rhetoric sentences to be classified as the target rhetoric category.

\section{Related Work}

\subsection{Emotion Neurons}

The exploration of emotion neurons aims to uncover the internal representation mechanisms of emotional information in LLMs, providing theoretical support for the fine-grained steering of emotion understanding and generation \cite{smith2011emotion,jayasinghe2025systematic,wang2025instructavatar,zheng2025stpar,zheng2025multi}.
Existing studies \cite{liao2025my,huangfu2025non} mostly focus on optimizing external emotion-related tasks.
With the advancement of interpretability research, the exploration of internal mechanisms at the neuronal level has gradually emerged as a research hotspot \cite{zhang2025exploring,tak-etal-2025-mechanistic}.
The concept of emotion neurons originates from 
\citet{radford2018learning}, referring to a set of neurons in the model that exhibit selective responses to specific emotions.
\citet{lee2025large} validate the existence of emotion neurons in Llama-series models, finding that their distribution varies with model scale.
\citet{di2025llamas} demonstrate that linear classifiers can achieve high-accuracy emotion recognition via probing techniques.

\subsection{Rhetorical Models}

Research on rhetorical processing revolves around two core tasks: recognition \cite{yang2025cultural,cocchieri2025you,zhang2025incongruity,ding2025zero,} and generation \cite{stowe2021metaphor,zhong2024let,goel2025target,zheng2024self}. 
In rhetorical recognition tasks, feature engineering and fine-tuning of pre-trained models are the mainstream technical approaches. 
\citet{saravia2018carer} achieve rhetoric type classification using contextual semantic features yet lack in-depth analysis of the underlying encoding mechanisms of rhetorical information in the model.
\citet{rajakumar2025evaluating} demonstrate the advantage of lightweight models in balancing efficiency and performance for practical deployment, but they do not provide an explanation for the internal operational principles of rhetorical representations.
In rhetorical generation tasks, 
\citet{deng2023rephrase} propose prompt reformulation to optimize the model’s comprehension of ambiguous queries. 
\citet{benara2024crafting} introduce a question-answering embedding approach that enables modeling of semantic representations associated with rhetorical expressions.

\section{Conclusion}

In this paper, we systematically explore emotion and rhetoric neurons in LLMs, addressing key gaps: 
inadequate exploration of rhetorical neurons, ambiguous emotion-rhetoric associations, and unreliable causal validation with traditional masking methods. 
Via a synergistic framework integrating neuron recognition, adaptive causal validation masking, and controllable causal intervention, we uncover their distribution and functional mechanisms, establishing a reliable system for causal validation and regulation. 
Experiments show our adaptive masking resolves counterintuitive drawbacks of traditional methods, yielding a robust tool for neuronal causal attribution.
Neuro modulation enables directed induction of emotional and rhetorical outputs and emotion-task performance gains via rhetoric neurons, laying a theoretical and technical foundation for fine-grained LLM steering.

\section*{Limitations}
Despite uncovering the internal neural mechanisms underlying emotion and rhetoric processing and advancing the interpretability of Large Language Models (LLMs), this study is not without limitations. 
First, the research scope is confined to 6 basic emotion categories and 4 core rhetorical devices; future work may extend to complex emotional states and diverse rhetorical forms. 
Second, neuronal manipulation in the current framework relies on static functional vectors, so context-aware dynamic adjustment strategies could further enhance manipulation precision.

\section*{Acknowledgments}

This work was supported by Ant Group through the CCF-Ant Research Fund. 
Fei Li and Donghong Ji are co-corresponding authors.

\bibliography{main}
\bibliographystyle{acl_natbib}

\end{CJK}
\end{document}